\documentclass[letterpaper, 10 pt, conference]{ieeeconf} 

\IEEEoverridecommandlockouts                

\usepackage{graphics} 
\usepackage{epsfig} 
\usepackage{mathptmx} 
\usepackage{times} 
\usepackage{amsmath} 
\newcommand\norm[1]{\left\lVert#1\right\rVert} 
\usepackage{amssymb, amsfonts}

\usepackage{color}
\usepackage{multirow}

\usepackage{algorithm}
\usepackage[noend]{algpseudocode}

\usepackage{nicefrac}   
\usepackage{microtype}  
\usepackage{booktabs,
            tabularx}
\usepackage{siunitx}

\usepackage{float}
\usepackage{multirow}
\usepackage{siunitx}
\usepackage{etoolbox}
\newrobustcmd{\B}{\fontseries{b}\selectfont} 

\usepackage{tikz}
\usetikzlibrary{shapes.geometric}
\usepackage{pgfplots}
\usetikzlibrary{positioning}
\usepackage{cancel}
\usepackage{subfig}
\usepackage{url}
\usepackage{cite}
\newtheorem{theorem}{Theorem}[section]

\usepackage{etoolbox}
\makeatletter
\patchcmd{\@makecaption}
  {\scshape}
  {}
  {}
  {}
\makeatother

\usepackage[
singlelinecheck=false 
]{caption}

\usepackage{mathtools}
\DeclarePairedDelimiter{\ceil}{\lceil}{\rceil}

\title{\LARGE \bf
EEG-GNN: Graph Neural Networks for Classification of Electroencephalogram (EEG) Signals
}

\author{Andac Demir$^{1}$, Toshiaki Koike-Akino$^{2}$, Ye Wang$^{2}$, Masaki Haruna$^{3}$, Deniz Erdogmus$^{1}$
\thanks{$^{1}$A. Demir, and D. Erdogmus are with Cognitive Systems Laboratory, Electrical and Computer Engineering Department, Northeastern University, Boston, MA 02115, USA (e-mail: demir.a@ece.neu.edu).} 
\thanks{
$^{2}$T. Koike-Akino and Y. Wang are with Mitsubishi Electric Research Laboratories (MERL), Cambridge, MA 02139, USA (e-mail: \{koike, yewang\}@merl.com).}
\thanks{$^{3}$M. Haruna is with Advanced Technology R\&D Center, Mitsubishi Electric Corporation (MELCO), Amagasaki, Hyogo, Japan.}
}

\begin{document}

\maketitle
\thispagestyle{empty}
\pagestyle{empty}

%%%%%%%%%%%%%%%%%%%%%%%%%%%%%%%%%%%%%%%%%%%%%%%%%%%%%%%%%%%%%%%%%%%%
\begin{abstract}
Convolutional neural networks (CNN) have been frequently used to extract subject-invariant features from electroencephalogram (EEG) for classification tasks. This approach holds the underlying assumption that electrodes are equidistant analogous to pixels of an image and hence fails to explore/exploit the complex functional neural connectivity between different electrode sites. We overcome this limitation by 
tailoring the concepts of convolution and pooling applied to 2D grid-like inputs for the functional network of electrode sites. Furthermore, we develop various graph neural network (GNN) models that project electrodes onto the nodes of a graph, where the node features are represented as EEG channel samples collected over a trial, and nodes can be connected by weighted/unweighted edges according to a flexible policy formulated by a neuroscientist. The empirical evaluations show that our proposed GNN-based framework outperforms standard CNN classifiers across ErrP, and RSVP datasets, as well as allowing neuroscientific interpretability and explainability to deep learning methods tailored to EEG related classification problems. Another practical advantage of our GNN-based framework is that it can be used in EEG channel selection, which is critical for reducing computational cost, and designing portable EEG headsets.   
\end{abstract}

\section{Introduction}

Deep convolutional neural networks (CNN) have been frequently used in extraction of task relevant features from physiological data, such as electroencephalogram (EEG)
and electromyogram (EMG) signals, to devise more robust human-machine interfaces (HMI). However a generic CNN is good at learning features from grid-like data structures like images, where each pixel is equidistant to neighboring pixels. Feeding EEG data to a CNN typically uses two methodologies:

\begin{enumerate}
    \item Applying 2D convolutions to each EEG trial, which is presented a pseudo-image $\mathbb{R}^{C \times T}$, where $C$ denotes number of EEG channels, and $T$ denotes number of discretized time samples, which effectively treats the EEG channels and time samples like spatial dimensions for CNN processing.
    \item Applying 1D convolutions along only the time axis of the EEG trial, while treating the EEG channels as separate channels of the CNN processing.
\end{enumerate}

In the 2D input case, arbitrarily stacking the time samples for each of the EEG channels into a single row ignores standard coordinates of EEG electrodes on a spherical head model. Neglecting functional neural connectivity between different parts of the brain oversimplifies the EEG feature extraction process. On the other hand,
CNNs using 1D convolutions with the EEG channels used as CNN channels underperform compared to the 2D convolutional case, since CNNs are only good at learning local spatial patterns, and not an effective approach to explore long-term temporal dependencies for not being sensitive to the order of timesteps.

The main contributions of this work over the existing studies are as follows: 
\begin{itemize}
   \item EEG-GNN properly maps the network of the brain as a graph, where each electrode used to collect EEG data according to intl. 10-5 system represents a node in the graph and time samples acquired from an electrode corresponds to that node's feature vector.
   \item Adjacency matrix of this graph can be constructed flexibly,
   e.g., i) every pair of nodes is connected by an unweighted edge, 
   ii) every pair of nodes is connected by an edge weighted by the functional neural connectivity factor, which is the Pearson correlation coefficient between the feature vectors of the two nodes, iii) a sparse adjacency matrix can be designed under the constraint only nodes that are closer than a heuristic distance are connected, or iv) a sparse adjacency matrix can be constructed via k-nearest neighbors (k-NNG).
   \item One of the major drawbacks to using CNNs to classify EEG data is they fail to provide a brain connectivity mapping by identifying Regions of Interests (ROIs), whereas EEG-GNN can learn and visualize the connectivity between salient nodes, which addresses a critical issue of neuroscientific interpretability.
\end{itemize}

\begin{figure}[t]
\centerline{\includegraphics[scale=0.78]{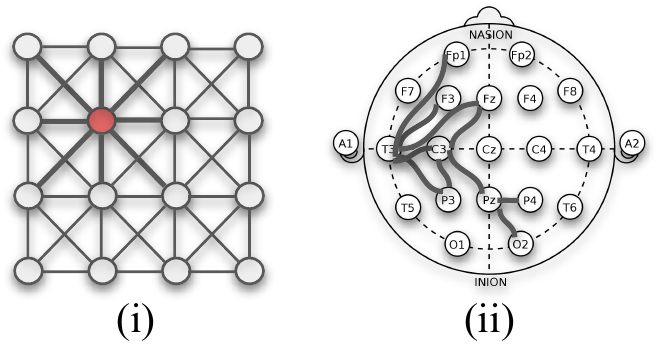}}
\caption{(i) is 2D CNN convolution, and the graph structure is analogous to pixels of an image. We know functional neural connectivity between electrode sites is arbitrary like in (ii). Cognitive activity of one hemisphere or lobe is more salient than the others for a specific EEG classification task, and there are intricate relationships between different electrode sites, which need to be explored.}
\label{fig:edge_index}
\end{figure}

\section{Preliminary to GNNs}
\begin{figure*}[t]
\centerline{\includegraphics[scale=0.7]{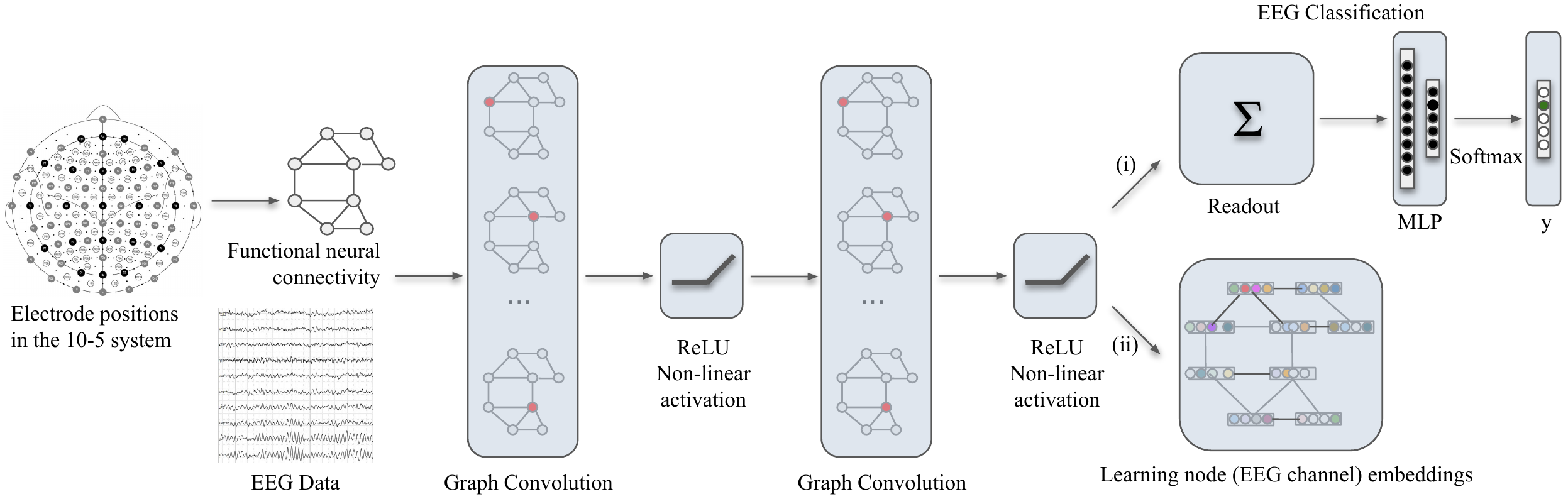}}
\caption{An EEG-GNN with graph convolutional operators and ReLU non-linearity applied on EEG signals mapped onto the graph structure of the functional neural connectivity between EEG electrode sites. Graph convolution layer is based on neighborhood aggregation approach, and encapsulates each EEG channel's hidden state vector by aggregating information from its neighboring electrode sites. As the number of graph convolution layers increases, we can get the hidden state vector of further neighborhoods. For 2 graph convolution layers, each node (EEG electrode site) aggregates information from its nearest neighbor and 2-hop neighbors. Following (i), we aggregate the node representations from the final iteration of graph convolution via \texttt{READOUT} function to learn the representation vector of the entire graph. Then, the graph representation vector is classified by a multi-layer perceptron (MLP) using softmax activation at the output layer. Optionally, following (ii), we transform the output from the final iteration of graph convolution with ReLU non-linearity, and learn an embedding for each EEG channel. This is particularly useful for tasks that require EEG channel selection instead of task classification.}
\label{fig:EEG-GNN classification}
\end{figure*}

A graph is a topological space which arises from a triplet $G=(V,E,W)$ of vertices, edges and weights. Vertices are simply a set of $n$ labels where $n$ is the total number of nodes: $V(G)=\{1,\dots,n\}$, where $n:=|V|$. Edges, $e_{ij}$,  are ordered pairs of labels $(i,j)$.  Weights, $w_{ij}$, are associated to edges. $w_{ij}$ represents strength of the influence of node $j$ on node $i$. 

The adjacency matrix of $G$ is the (typically sparse) matrix $\mathbf{A}$ with entries given by $\mathbf{A}_{ij}=w_{ij}$ for all $i,j$. If $G$ is symmetric, then $\mathbf{A}$ is symmetric: $\mathbf{A}=\mathbf{A}^\mathbf{T}$. For the particular case in which $G$ is unweighted, weights are interpreted as units: $\mathbf{A}_{ij}=1$, if node $i$ and node $j$ are connected, and otherwise 0. In an unweighted graph, $\mathbf{A}$ has the same sparsity pattern, except that now all nonzero entries will be 1. 

The degree of a node $i$, denoted as $d_{i}$, is the sum of weights in all of the incident edges that connect $i$ from its neighbors.
The degrees of all nodes are grouped into the degree matrix $\mathbf{D}$. This is a diagonal matrix whose diagonal entry is the $d_{i}$. $\mathbf{D}$ can also be written in terms of the adjacency matrix, because the diagonal of $\mathbf{D}$ corresponds to the sum of the rows of $\mathbf{A}$.

Having defined adjacency and degree matrices, we can now define the Laplacian matrix, $\mathbf{L}$, of a graph,
\begin{equation}
    \mathbf{L}=\mathbf{D}-\mathbf{A}
\end{equation}
The Laplacian can also be written explicitly in terms of the weights of the graph. Since $\mathbf{D}$ is diagonal, off-diagonal entries are simply given by opposite values of the corresponding entries of $\mathbf{A}$. $\mathbf{L}_{ij} = -\mathbf{A}_{ij} = -w_{ij}$. Assuming the graph has no self loops, the diagonal entries of $\mathbf{A}$ are null, hence the diagonal entries of $\mathbf{L}$ are simply the diagonal entries of $\mathbf{D}$: $\mathbf{L}_{ij}=d_{i}$.

Normalized versions of $\mathbf{A}$ and $\mathbf{L}$ are also utilized as matrix representations of graphs. Both normalized adjacency and Laplacian are defined by pre-emposed multiplication by the inverse of square root of $\mathbf{D}$,
\begin{equation}
    \tilde{\mathbf{A}} = \mathbf{D}^{-1/2}\mathbf{A}\mathbf{D}^{1/2} \quad \text{and} \quad \tilde{\mathbf{L}} = \mathbf{D}^{-1/2}\mathbf{L}\mathbf{D}^{1/2}
\end{equation}
These pre-emposed multiplications by $\mathbf{D}^{-1/2}$ result in representations in which weights are expressed relative to the degrees of individual nodes. Both $\tilde{\mathbf{A}}$ and $\tilde{\mathbf{L}}$ will be symmetric if the graph is symmetric. Further observed, given these definitions $\tilde{\mathbf{L}}$ can also be written as, 
\begin{equation}
    \tilde{\mathbf{L}} = \mathbf{I}
     - \tilde{\mathbf{A}}
\end{equation}
\paragraph{Graph Shift Operators}
Graph shift operator $\mathbf{S}$ is a stand in operator for any of the matrix representations of the graph. We can set it $\mathbf{S}=\mathbf{A}$, $\mathbf{S}=\mathbf{L}$, $\mathbf{S}=\tilde{\mathbf{A}}$ or $\mathbf{S}=\tilde{\mathbf{L}}$. The specific choice of $\mathbf{S}$ matters in practice, but most of results and analysis hold for any choice of $\mathbf{S}$.

\section{Methodology}
The path towards scalable machine learning on graphs begins from the generalization of the convolutional operator in CNNs to signals supported on graphs. Once the convolutional operator is generalized to graphs, we can easily generate graph filter banks, combine these graph filter banks with pointwise non-linearities, and then stack them into layers to create GNNs. Graph convolutions follow a neighborhood aggregation strategy collecting the node features, $\mathbf{X} \in \mathbb{R}^{|V|\times F}$ (where $|V|$ denotes number of nodes, and $F$ denotes the size of a node feature vector) within each node's K-hop neighbors to learn a representation vector of a node, $h_{v}$, or the entire graph, $h_{G}$. 
\begin{equation}
    h_{G} = \sum_{k=0}^{K-1}\sigma(\mathbf{S}^{k}\mathbf{X}\mathbf{W}_{k}),
\end{equation}
where $\sigma$ is the pointwise non-linearity, $\mathbf{S}\in \mathbb{R}^{|V|x|V|}$ is the graph shift operator, and $\mathbf{W}_{k}\in \mathbb{R}^{F\times G}$ is the graph filter, where $G$ denotes the number of filters applied to each node feature vector. Multiplication by $\mathbf{W}$ gives the linear combination of features at each node. $h_{G}$ iteratively updated $K$ times to capture structural information from K-hop neighbors. We compare the discriminative power of various different GNN implementations for classifying EEG, and benchmark their performance against the state-of-the-art CNN models.

\paragraph{\textbf{GraphSAGE}}
GraphSAGE learns an aggregated neighborhood embedding, $h_{N(v)}^k$,  via $\texttt{AGGREGATE}_{k}$ functions, which distill high dimensional information from a node's K-hop neighbors, $h_{u}^{k-1}, \ \forall u \in N(v)$, in which $N(v)$ is the set of node $v$'s  K-hop neighbors and $u$ denotes the nodes uniformly sampled from $N(v)$~\cite{hamilton2017inductive}. The aggregated neighborhood embedding is then concatenated with the current node representation, $h_{v}^{k-1}$, and modulated by a trainable weight matrix (by feeding through a fully connected layer), $W_{k}$, which learns to propagate different levels of information for different search depths, $k$, of a node, and incorporates information about the graph structure. The output is passed through a ReLU non-linearity, $\sigma$. Algorithm \ref{alg:graphsage} summarizes the procedure for the GraphSAGE algorithm.   

\begin{algorithm}[t]
\caption{GraphSAGE forward propagation - learning graph embedding, $h_{G}$.}
\begin{algorithmic}[1]
\State $h^{0}_{v}\leftarrow \mathbf{X}_{v}, \forall v \in V$\; Initialize a representation vector for each node
\For{$k=1\dots K$}
    \For{$v \in V$}
        \State $h^{k}_{N(v)}\leftarrow \texttt{AGGREGATE}_{k}(\{h_{u}^{k-1},\ \forall u \in N(v)\})$
        \State $h^{k}_{v}\leftarrow \sigma\left(\mathbf{W}^{k}\cdot \texttt{CONCAT}(h_{v}^{k-1},\ h_{N(v)}^{k})\right)$
    \EndFor
    \State $h_{v}^{k}\leftarrow h_{v}^{k}/\norm{h_{v}^{k}}_{2}$
\EndFor
\State $h_{G}\leftarrow \texttt{READOUT}(\{h_{v}^{K}, \ \forall v\in V\})$
\end{algorithmic}
\label{alg:graphsage}
\end{algorithm}

\texttt{AGGREGATE} and \texttt{READOUT} functions must be invariant to permutation of input node representations such as summation or graph level max/mean pooling. We apply an element-wise mean-pooling as \texttt{AGGREGATE}, where each K-hop neighbor's representation vector is fed through a fully connected layer with ReLU non-linearity, and then an element-wise mean-pooling operation is applied to each of the computed node representations. \texttt{READOUT} returns the sum of node representations after graph convolution at the final iteration.
\begin{align}
    \texttt{AGGREGATE}_{k}=\texttt{mean}(\{\sigma(\mathbf{W}_{\text{pool}}h_{u_{i}}^{k}+b), \ \forall u_{i} \in N(v)\})
    \label{eq:aggregate}
\end{align}

\paragraph{\textbf{Graph Isomorphism Network (GIN)}}
GIN is conceptually inspired by Weisfeiler-Lehman (WL) isomorphism test, which reduces graphs to their canonical forms to check whether they are topologically identical. Graph isomorphism requires there is a bijective function $f: G_{1} \rightarrow G_{2}$ mapping a graph $G_{1}$ to another graph $G_{2}$, while preserving adjacencies. GIN proposes that if $G_{1}$ and $G_{2}$ are non-isomorphic, then their embeddings $h_{G_{1}}$ and $h_{G_{2}}$ cannot be identical~\cite{xu2018powerful, kim2020understanding}. It also proves if \texttt{AGGREGATE} and \texttt{READOUT} functions are both permutation invariant and injective, then GNN is at most as powerful as WL isomorphism test at recognizing different graph structures~\cite{xu2018powerful}.

\begin{figure}[t]
\centerline{\includegraphics[scale=0.8]{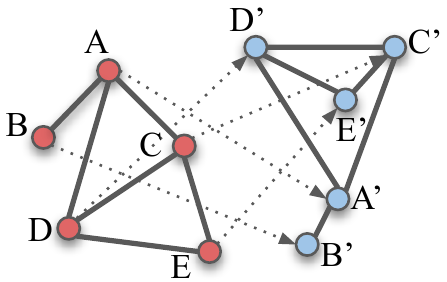}}
\caption{An injective surjective mapping (bijection) between 2 topologically identical graphs. Adjacencies are preserved.}
\label{fig:isomorphism}
\end{figure}

\begin{figure}[t]
\centerline{\includegraphics[scale=0.75]{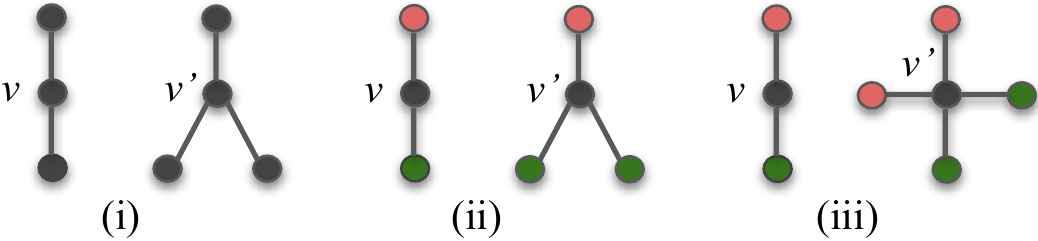}}
\caption{(i) $v$ and $v'$ would have the same node embedding, if \texttt{AGGREGATE} function in (i) is \texttt{mean} or \texttt{max}, in (ii) is \texttt{max}, and in (iii) is \texttt{mean} or \texttt{max}.}
\label{fig:wl_test}
\end{figure}

Nodes in different graph structures get the same embedding, when \texttt{mean} and \texttt{max} operators are used to aggregate information from neighboring nodes, as illustrated in Fig. \ref{fig:wl_test}, because these operators are non-injective. GNNs can only be as powerful as WL isomorphism test, if and only if \texttt{AGGREGATE} function maps an identical embedding for two nodes from two graphs, only when these two nodes have same subtree structures and same feature vectors in neighboring nodes~\cite{wu2020comprehensive}. According to universal approximation theorem, an MLP with one hidden layer can approximate any measurable functions. Hence, GIN approximates the solution to formulate an injective and permutation invariant aggregation operator via training an MLP with a single hidden layer as revealed in Equation \ref{eq:gin_equations}, where $\lambda$ is a learnable parameter. Graph representation concatenates the summations of node embeddings at the same iteration.

\begin{figure}[t]
\centerline{\includegraphics[scale=0.75]{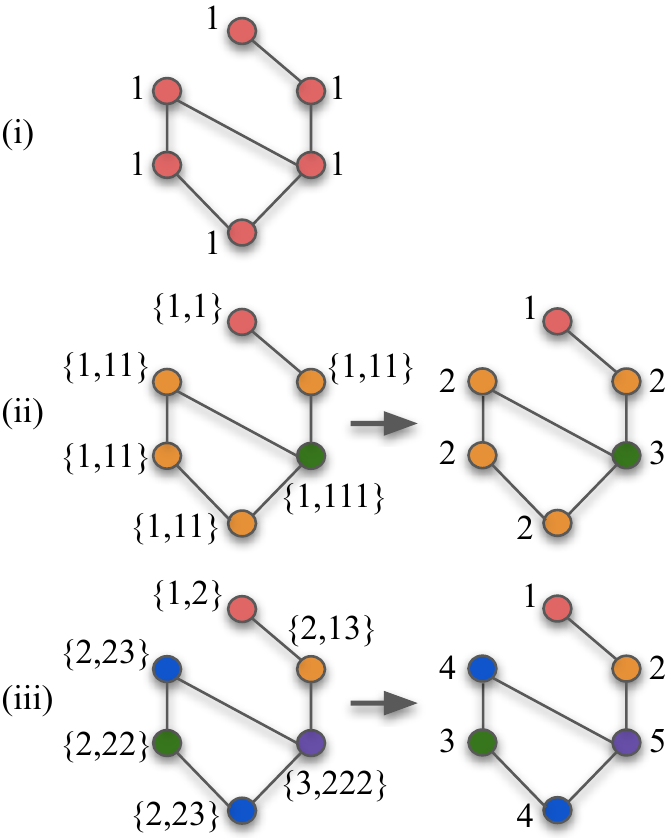}}
\caption{\textbf{WL Subtree Kernel Method} (i) All vertices are assigned the initial color label 1, (ii) A signature string is constructed for each vertex by concatenating its own color label and its neighbors' color labels. Then, all the signature strings are compressed into their new integer color labels in the lexicographical order, (iii) This process is repeated until convergence to produce a canonical form of the graph. Vertices with the same color label have structurally similar roles in the graph.}
\label{fig:wl_test_procedure}
\end{figure}

\begin{theorem}[Universal Approximation Theorem]
\label{Universal Approximation Theorem}
Let $\sigma$ be a non-linear activation function, and $x\in I_{n}$. Given the output of a one hidden layer MLP with weights $w_{j}, \alpha_{j}$ and biases $b_{j}$ of the form,
\[G(x)=\sum_{j=1}^{N}\alpha_{j}\sigma(w_{j}x + b)\]
Then for any $f\in C(I_{n})$, and $\epsilon>0$, there is a $G(x)$ such that 
\[|G(x)-f(x)|\leq \epsilon, \quad \forall x \in I_{n}.\]
\end{theorem}

\begin{equation}
    \begin{split}
        & h_{v}^{k} \leftarrow \texttt{MLP}^{k}\left((1+\lambda^{k})\cdot h_{v}^{k-1} + \sum_{u}h_{u}^{k-1}, \forall u \in N(v)\right) \\
        & h_{G}\leftarrow \texttt{CONCAT}\left(\texttt{READOUT}(\{h_{v}^{k},\  \forall v\in V\})\ |\ k=0,1\dots K\right)
    \end{split}
    \label{eq:gin_equations}    
\end{equation}

Since composition of injective functions is also injective, node representations can be evaluated with a single MLP,
\begin{equation}
    h_{v} \leftarrow \texttt{MLP}\left((1+\lambda^{k})\cdot h_{v}^{k-1} + \sum_{u}h_{u}^{k-1}, \forall u \in N(v)\right)
    \label{eq:single_mlp_gin} 
\end{equation}

\begin{theorem}
    Let $f:G\rightarrow \mathbb{R}^{d}$ be a GNN, and $G_{1}, G_{2}$ represent 2 graphs. $f$ is as powerful as the WL isomorphism test if \texttt{AGGREGATE} and \texttt{READOUT} functions are injective. Since \texttt{READOUT} projects distinct node representations in the domain to distinct graph representations in the codomain, we only need to prove aggregating neighborhood features must be injective~\cite{xu2018powerful}. Given node representations $h_{v}^{k}$ derived by injective functions $\psi$ and $\phi$, and node labels $l_{v}^{k}$ assigned by WL isomorphism test operator with injective function $\varphi$, \[h_{v}^{k} \leftarrow \psi\left(h_{v}^{k-1}, \ \phi \left(\Big\{h_{u}^{k-1}, \forall u \in N(v)\Big\}\right)\right)\] 
    \[l_{v}^{k} \leftarrow \ \varphi\left( l_{v}^{k-1}, \Big\{l_{u}^{k-1}, \forall u \in N(v)\Big\}\right)\]
    We can prove by induction, $\exists g\in \{ g(a)=g(b)\implies a=b, \ \forall a,b\in \mathbb{R}\}$ that satisfies $h_{v}^{k}=g(l_{v}^{k})$.\\
    \underline{Base case} for $k=0$: $h_{v}^{0}=g(l_{v}^{0}), \ \forall v\in G_{1}, G_{2}$, because initial node representations are identical for $f$ and WL subtree kernel method.\\ \underline{Inductive step} for $k=k-1$: 
    \[h_{v}^{k} \leftarrow \psi\left( g\left(l_{v}^{k-1}\right), \ \phi \left(\Big\{g\left(l_{u}^{k-1}\right), \forall u \in N(v)\Big\}\right)\right)\] 
    Since the composition of injective functions is injective, this can be rewritten using another injective function $\omega$,
    \[h_{v}^{k} \leftarrow \omega\left( l_{v}^{k-1}, \ \Big\{l_{u}^{k-1}, \forall u \in N(v)\Big\}\right)\] 
    which is identical to,
    \[h_{v}^{k} \leftarrow \omega \circ \varphi^{-1}\varphi\left(l_{v}^{k-1}, \ \Big\{l_{u}^{k-1}, \forall u \in N(v)\Big\}\right) = \omega \circ \varphi^{-1}\left(l_{v}^{k}\right)\]
    Since the composition $\omega \circ \varphi^{-1}$ is also injective, $f$ is as powerful as WL isomorphism test to decide the multisets $\Big\{l_{v}^{k}\Big\}$ are unique for non-isomorphic $G_{1}$ and $G_{2}$ at each iteration $k$.
\end{theorem}

Pooling in CNNs downsamples the number of features in the input, which reduces the number of model parameters, and helps to avoid overfitting. SortPool, EdgePool, SagPool and Set2Set propose different pooling operations that play a similar role in GNNs. 

\paragraph{\textbf{SortPool}}
The key insight here is sorting vertices based on their structural importance established by WL isomorphism test in order to sequentially extract features from a graph in a consistent order~\cite{zhang2018end}. After $K$ iterations of graph convolution, SortPool layer gets an input of size $|V|\times K\cdot|h_{v}|$, where $|V|$ denotes the number of nodes, and $|h_{v}|$ denotes the size of a node's embedding vector, which is assumed to be identical for all the nodes. The output of SortPool layer has size $\rho\times K\cdot|h_{v}|$, where $\rho$ is a heuristically selected parameter, and $\rho<|V|$. SortPool layer sorts the color labels in descending lexicographical order according to WL subtree kernel method, and chooses the first $\rho$ number of nodes. Then, the sorted output is reshaped as a row vector of size $\rho\cdot K\cdot|h_{v}| \times 1$ and applied 1D convolutional layers with kernel size $K\cdot|h_{v}|$ followed by MaxPool. Final MaxPool output is fed through a dense layer with softmax activation.

\paragraph{\textbf{EdgePool}}
Given a weighted graph, softmax function is applied to all edge weights to compute edge scores, denoted as $s_{ij}$. According to these scores, edges are iteratively contracted unless nodes have already been part of a contracted edge~\cite{diehl2019towards, diehl2019edge}. Edges between contracted nodes are preserved. If there is an isolated node (a node with degree 0) after edge contraction, edges between the isolated node and contracted nodes are reconstructed. In the process of edge contraction, node features are combined, and then multiplied by the edge score, which allows the gradient to backpropagate through edge scores. Let $e=\{v_{i}, v_{j}\}$ is the edge contracting 2 nodes,
then $h_{v_{ij}} = s_{ij}(h_{v_{i}} + h_{v_{j}})$.

\begin{figure}[t]
\centerline{\includegraphics[scale=0.75]{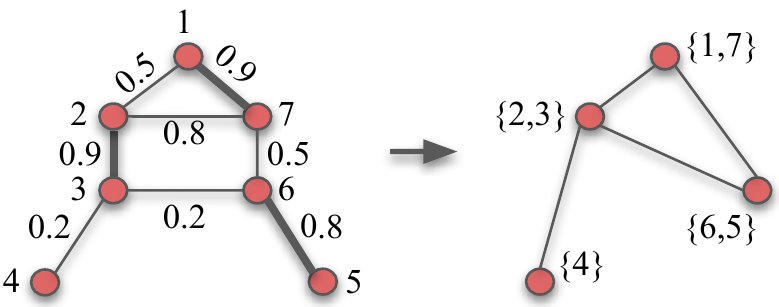}}
\caption{After EdgePool operator, number of nodes is downsampled by a fixed ratio of 2.}
\label{fig:edgepooling}
\end{figure}

\begin{figure}[t]
\centerline{\includegraphics[scale=0.70]{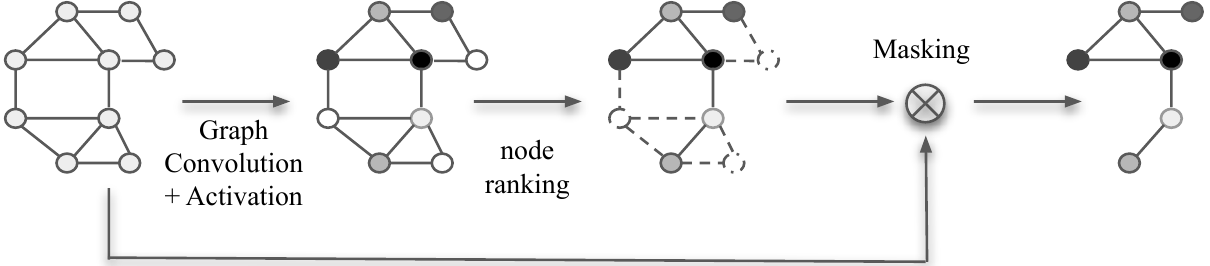}}
\caption{Downsampling with SagPool operator.}
\label{fig:sagpooling}
\end{figure}

\paragraph{\textbf{SagPool}}
SagPool uses both node features and graph topology for pooling via a masking operator~\cite{lee2019self}, as illustrated in Figure~\ref{fig:sagpooling}. First, the graph convolution layer computes self-attention scores, $Z\in \mathbb{R}^{|V|\times 1}$,
\begin{equation}
    Z = \sigma(\mathbf{S}\mathbf{X}\mathbf{W}_{att})
\end{equation}
where $\mathbf{W}_{att}\in \mathbb{R}^{F\times 1}$ denotes the learnable pooling parameters. Top-rank function~\cite{gao2019graph} selects the indices of most useful $\rho\times|V|$ number of nodes according to $Z$ for a heuristic parameter $\rho\in(0,1]$. The pooled output is the element-wise multiplication of selected node features and attention scores, 
\begin{equation}
    \begin{split}
        &\text{idx} = \text{top-rank}\Big(Z, \ceil[\big]{\rho|V|}\Big), \quad Z_{\text{mask}} = Z_{\text{idx}} \\
        & X_{\text{out}} = X_{\text{idx}}\odot Z_{\text{mask}}
    \end{split}
\end{equation}

\paragraph{\textbf{Set2Set}}It's an extension of Seq2Seq framework, which uses LSTM's with attention mechanism for graph level pooling. The pooled output, denoted as $q_{t}^{*}$, is the concatenation of the weighted sum of node embeddings, and the query vector, which reads from the memory of an LSTM~\cite{vinyals2015order}.  

\begin{algorithm}[t]
\caption{Set2Set Pooling Operator}
\begin{algorithmic}[1]
\State $q_{0}^{*}\leftarrow \textbf{zeros}(1, \text{input dim.})$\; Initialize a zero vector for the query vector $q$
\State $\text{hidden}_{0}\leftarrow \textbf{zeros}(1, \text{output dim.})$\; Initialize a zero vector for the hidden state of LSTM cell
\State $\text{cell}_{0}\leftarrow \textbf{zeros}(1, \text{output dim.})$\; Initialize a zero vector for the cell state of LSTM cell
\State $h_{i}$\; Embedding of node $i$
\For{$i \in \{1,2,\dots|V|\}$}
    \State $q_{t}, \text{hidden}_{t}, \text{cell}_{t} = \textbf{LSTM}(q^{*}_{t-1}, \text{hidden}_{t-1}, \text{cell}_{t-1})$
    \State
     $\alpha_{i}=\frac{h_{i}^Tq_{i}}{\sum_{i}h_{j}^Tq_{t}}$\; (attention params.)
    \State $r_{t}=\sum_{i}\alpha_{i}h_{i}$\; (attention readout)
    \State $q_{t}^{*}=[q_{t}, r_{t}]$
\EndFor
\State $h_{G}=\textbf{ReLU}(\textbf{Dense}(q_{t}^{*}))$
\end{algorithmic}
\label{alg:set2set}
\end{algorithm}

\section{Datasets \& Software}
\paragraph{\textbf{ErrP}} 
The detection of error-related potentials (ErrP) to improve the accuracy of P300-based BCI speller~\cite{margaux2012objective}.\footnote{\url{https://www.kaggle.com/c/inria-bci-challenge/}}
The dataset was recorded from $16$ healthy subjects participating in an offline P300 spelling task. Spelling task had a fast mode (each item was flashed 4 times), and a slow mode (each item was flashed 8 times). Each subject performed $340$ trials. If there is an inconsistency between subject's intention and BCI system, elicited ErrP should be detected. 
EEG data were recorded at a downsampled rate of $200$~Hz from $56$ channels. A trial has $250$ discretized time samples, and is associated with a binary class label: erroneous (inferred item is different from the intent of subject) or correct feedback.

\paragraph{\textbf{RSVP}}
A BCI system to type based on rapid serial visual presentation (RSVP) paradigm~\cite{orhan2012rsvp}.\footnote{\url{http://hdl.handle.net/2047/D20294523}} The dataset was collected from $10$ healthy subjects, and consists of $41{,}400$ trials of $16$ channel EEG data. A g.USBamp biosignal amplifier with active electrodes was used to record trials during RSVP keyboard operations. Each trial has $128$ discretized time samples, and is associated with one of the 4 labels: emotion elicitation, resting-state, or motor imagery/execution task.

We use PyTorch Geometric v1.8.0~\cite{Fey/Lenssen/2019} to implement GNN variants. All models were trained with a minibatch size of 256 for 400 epochs on NVIDIA Tesla K80 12GB GPU, and optimized by Adam with an initial learning rate of 0.001, which decays into half every 50 epochs. EEG trials from all subjects were shuffled, first $80\%$ of EEG trials was used for training, and the last $20\%$ was used for validation. If there was an improvement in classification accuracy on the validation dataset, model checkpoints were saved.

\section{Results \& Discussion}

\paragraph{\textbf{Edge Formation}} We present 6 different methods to connect electrode sites by edges e.g., i) each pair of electrode sites is connected (complete graph), ii) graph is complete, and allows self-loops, iii) electrode sites $x$ and $y$ are connected by an edge, if the distance between is among the k-th lowest distance that connects an electrode to $x$ (k-NNG), iv) k-NNG allows self-loops, v) each pair of electrode site is connected, if the distance between is lower than a heuristic distance threshold, vi) distance thresholding for the edge formation procedure allows self-loops. Empirical results in Table~\ref{tab:edge} and~\ref{tab:dist} indicate that connecting electrode sites with only their nearest neighbor provides a classification accuracy as good as a complete graph, while also saving training time, and model size. Therefore, we conclude that using fewer electrodes, and a sparse adjacency matrix does not cause a drop in classification performance, but has a positive impact on the system efficiency.   
\paragraph{\textbf{Parameter Regularization}} Although GNNs have a high expressive power learning over graph structured data, there is no theoretical guarantee on their generalizability. Empirical results show that they are prone to overfitting for small datasets, and suffer from issues of poor convergence. In our experiments, we compare the performance of L1, L2, and elastic net regularization across different values of hyperparameters to avoid overfitting. While the L1 penalty imposes sparsity by adding the absolute value of the magnitude of model weights modulated by the hyperparameter $\alpha$ to the loss objective, L2 penalty imposes feature selection, and shrinks weights towards 0, by adding the square of the magnitude of weights modulated by $\beta$. Elastic net performs concurrent regularization with both the L1 and L2 penalty. In Table~\ref{tab:res}, we finetune $\alpha$ and $\beta$, and report classification performance for neural connectivity between electrode sites constructed according to k-NNG $(k=1)$. Fast convergence, and sub-optimal optimization of model parameters were common issues (as also reported by other practitioners training GNNs in different domains~\cite{mercado2021graph}), despite our tests across smaller learning rates, parameter regularization, data augmentation, and node feature selection to reduce model bias, and variance. GNN models have one pitfall in that they tend to overfit quite heavily in small datasets.
\paragraph{\textbf{Data Augmentation}} 
Training data is augmented by adding white Gaussian noise (AWGN) to every channel of the original
signal. The variance of this zero-mean Gaussian random variable affects the average noise power, and is specified to obtain SNR levels of 10, 5, and 2 dB. Desired SNR level is the difference between signal power and noise power in dB. This provides a three-fold increase in the number of EEG trials in the training datasets.
\paragraph{\textbf{Temporal Compression}}
GNN models that classify EEG datasets defined over spatial-temporal graphs become more susceptible to overfitting as the number of features associated with each node increases due to a higher number of model parameters~\cite{wu2020comprehensive}. This turns out to be even more problematic while learning over EEG datasets, because number of samples per trial, which is equivalent to number of node features, is 250 for ErrP, and 128 for RSVP. We apply 1D convolutions with kernel size of $1\times3$ and stride 2 along the time axis of a trial, and then adopt batch normalization. This downsamples number of node features by 2, and also helps to capture temporal dependencies. We stack several layers of 1D convolutions, and reduce the number of node features to 32. 

\begin{table*}[t]
\centering
\caption{Task classification performance of GNN compared to a CNN classifier, AutoBayes models, and the ensemble of AutoBayes.}
\label{tab:bench}

\scriptsize
\begin{tabular}{c c c c c}
\toprule
Method & \multicolumn{2}{c}{ErrP} & \multicolumn{2}{c}{RSVP} \\
\cmidrule(lr){2-3} \cmidrule(lr){4-5}  
& Acc. & \# Model Params. & Acc. & \# Model Params.\\

\midrule
\multirow{1}{*}{EEG-GNN}
& \textcolor{red}{\boldmath$76.73\pm0.40$}   
& $106{,}562$
& \textcolor{red}{\boldmath$93.49\pm0.10$}  
& $83{,}138$ \\

\multirow{1}{*}{Standard CNN}
& $74.72\pm0.31$   
& $127{,}335$
& $93.07\pm0.15$  
& $268{,}865$\\

\multirow{1}{*}{Best of AutoBayes}
& $75.91\pm0.44$   
& $3{,}407{,}390$
& $93.42\pm0.15$  
& $2{,}005{,}917$\\

\bottomrule
\end{tabular}
\end{table*}

We benchmark the performance of EEG-GNN against the state-of-the-art CNN classifiers in Table~\ref{tab:bench}. Compared to a standard CNN classifier of similar model size, GNN models significantly improve the accuracy by $2.0\%$ for ErrP, whereas the improvement for RSVP is $0.4\%$. Compared to AutoBayes classifiers~\cite{demir2021autobayes}, which detect the conditional relationship between data features, task labels, nuisance variation labels (subject IDs), and potential latent variables in DNN architectures to identify the best inference strategy, GNN models still have higher classification performance, while reducing model size more than 20x. Although GNN models don't take advantage from adversarial learning using variations in subject IDs, they perform $0.8\%$ better for ErrP, and nearly same for RSVP. Identification of best Bayesian network trained in an adversarial setting helps with getting a non-dispersive distribution of the accuracies across different subjects and hence provides more robustness against subject variation; however GNN classifiers demonstrate an accuracy equivalent to the best Bayesian network selected by AutoBayes framework.

We further benchmark the performance of EEG-GNN against the standard CNN classifiers by exploring different network configurations in Fig.~\ref{fig:acc_vs_space complexity}. Pareto front connects the model configurations with superior performance proceeding from low to high complexity regime. It is observed that EEG-GNN performs significantly higher in low complexity regime than the standard CNN classifier. However, the performance gap reduces in high complexity regime. These results highlight the problem that scaling up GNN models to exploit multi-hop neighborhoods does not necessarily improve the performance. Vanishing gradient problem is encountered while training deeper GNN models. Using skip connections between graph convolutional layers can mitigate the effects of vanishing gradients, and allow to capture the representations of higher order graphs.

\section{Conclusion \& Future Work}

In this paper, we presented several GNN models along with various regularization strategies to model the functional neural connectivity between EEG electrode sites, and demonstrated GNN models outperform CNN models of different size and inference strategies in classification tasks across ErrP and RSVP datasets. There are many interesting directions for prospective research. For instance, we are currently learning models over unweighted graphs, but we can also use the Pearson correlation coefficient between EEG channels to represent the edge weights. Another interesting approach is to entirely eliminate a hand-engineered design of adjacency matrix, and learn a graph shift operator matrix along with model weights during training, in order to better capture the data topology. This would require parameterizing the graph shift operator matrix, transferring it to graph convolutional operators, and then appropriately taking gradients in backpropagation. 

\begin{figure*}[t]
\centering
\subfloat[ErrP Dataset]{
\scalebox{0.78}{\begin{tikzpicture}
\def\axisdefaultwidth{0.5\linewidth}
\def\axisdefaultheight{0.25\linewidth}
\pgfplotsset{every axis/.style={scale only axis}}
\begin{axis}[
    ylabel near ticks,
    xlabel near ticks,
    scaled y ticks = false,
    yticklabel style={/pgf/number format/fixed},
    ylabel={Accuracy (\%)},
    xmode=log,
    xlabel={Number of Parameters},
    xmin=1e3, xmax=1e6, ymin=50, ymax=100,
    legend style={nodes={scale=0.7, transform shape},draw=none},
    legend pos=south east,
    legend cell align={left},
    reverse legend
    ]
    \addplot[
        scatter/classes={b={blue}, c={red}},
        scatter, mark=*, only marks, 
        scatter src=explicit symbolic,
        visualization depends on={value \thisrow{label} \as \Label} 
    ] table [meta=class] {
    
        x y class label
        10935 63.62 b B1
        106562 77.99 c C1
        
        47325 71.08 b B2
        65730 79.65 c C2
        
        27665 69.26 b B3
        33090 77.62 c C3 
        
        54305 72.81 b B4
        143938 78.91 c C4

        127335 74.72 b B5
        275650 76.52 c C5

        226605 74.26 b B6
        41538 77.44 c C6

        180835 72.98 b B7
        477250 79.00 c C7

        150115 71.31 b B8
        49986 78.81 c C8

        102795 72.66 b B9
        99010 77.99 c C9

        478875 72.55 b B10
        831042 78.36 c C10

        238145 73.01 b B11
        58434 78.17 c C11

        255455 74.08 b B12
        132290 79.74 c C12

        129735 72.36 b B13
        21264 77.89 c C13
        
        127995 73.22 b B14
        3012 75.60 c C14
        
        1337 51.15 c C15
    };
    \legend{Standard CNN Classifier, EEG-GNN}
    
    \addplot[color=blue, mark=*] coordinates {
        (10935, 63.62) 
        (27665, 69.26) 
        (47325, 71.08) 
        (54305, 72.81) 
        (127335, 74.72)
    }; 
   
    \addplot[color=red, mark=*] coordinates {
        (1337, 51.15)  
        (3012, 75.60)  
        (21264, 77.89) 
        (49986, 78.81) 
        (65730, 79.65) 
        (132290, 79.74)
    }; 
\end{axis}
\end{tikzpicture}}
}
\hfill
\subfloat[RSVP Dataset]{
\scalebox{0.78}{\begin{tikzpicture}
\def\axisdefaultwidth{0.5\linewidth}
\def\axisdefaultheight{0.25\linewidth}
\pgfplotsset{every axis/.style={scale only axis}}
\begin{axis}[
    ylabel near ticks,
    xlabel near ticks,
    scaled y ticks = false,
    yticklabel style={/pgf/number format/fixed},
    ylabel={Accuracy (\%)},
    xmode=log,
    xlabel={Number of Parameters},
    xmin=0.5e3, xmax=5e6, ymin=67, ymax=100,
    legend style={nodes={scale=0.7, transform shape},draw=none},
    legend pos=south east,
    legend cell align={left},
    reverse legend
    ]
    \addplot[
        scatter/classes={b={blue}, c={red}},
        scatter, mark=*, only marks, 
        scatter src=explicit symbolic,
        visualization depends on={value \thisrow{label} \as \Label} 
    ] table [meta=class] {
        
        x y class label
        2005917 93.07 b B1
        25282 93.16 c C1
        
        3407390 92.23 b B2
        120514 93.02 c C2
        
        1703695 92.85 b B3
        83138 92.76 c C3 
        
        851847 91.90 b B4
        24418 93.50 c C4

        243385 90.22 b B5
        33730 93.27 c C5

        100989 89.81 b B6
        50114 93.14 c C6

        67326 88.37 b B7
        42178 92.64 c C7

        572274 87.70 b B8
        438210 92.64 c C8

        457819 91.85 b B9
        727 72.73 c C9

        268865 93.07 b B10
        792002 89.96 c C10

        763032 91.50 b B11
        1792 83.13 c C11

        508565 90.92 b B12
        13456 92.37 c C12

        42380 84.28 b B13
        116674 92.67 c C13
        
        19263 81.77 b B14
        50626 92.45 c C14
        
        635707 91.72 b B15
        83394 92.75 c C15
};
    \legend{Standard CNN Classifier, EEG-GNN}
    
    \addplot[color=blue, mark=*] coordinates {
        (19263, 81.77)  
        (42380, 84.28)  
        (67326, 88.37)  
        (100989, 89.81) 
        (268865, 93.07) 
        (2005917, 93.07)
    }; 
   
    \addplot[color=red, mark=*] coordinates {
        (727, 72.73)   
        (1792, 83.13)  
        (13456, 92.37) 
        (24418, 93.50) 
    }; 
\end{axis}
\end{tikzpicture}}
}
\caption{Accuracy vs. Space Complexity}
\label{fig:acc_vs_space complexity}
\end{figure*}

\begin{table*}[t]
\centering
\caption{Performance of datasets: Edge index matrix construction using a complete graph (all), a complete graph containing self-loops (all with self-loops), computing graph edges to the nearest k neighbors (k-NNG), and  k-NNG containing self-loops (k-NNG w. self-loops).}
\label{tab:edge}

\scriptsize
\begin{tabular}{c c c c c c c c c c}
\toprule
Dataset & Model &
\multicolumn{1}{c}{All} & \multicolumn{1}{c}{All w. Self-Loops} & \multicolumn{3}{c}{k-NNG} & 
\multicolumn{3}{c}{k-NNG w. Self-Loops}\\
\cmidrule(lr){5-7} \cmidrule(lr){8-10}  
& & & & k=1 & k=2 & k=4 & k=1 & k=2 & k=4\\

\toprule
\multirow{6}{*}{ErrP}
& GraphSage  & $74.44 \pm 0.75$   & $75.94 \pm 1.42$        & $74.04 \pm 0.98$        & $74.89 \pm 1.88$         &  $75.29 \pm 0.70$  &
$74.34 \pm 0.62$  & $74.47 \pm 0.88$ & \textcolor{red}{\boldmath$76.33 \pm 0.69$}\\
\cline{2-10}

& Set2Set    & $75.38 \pm 0.54$   & $74.62 \pm 0.17$        & $75.66 \pm 0.82$        & $74.37 \pm 0.83$        & \textcolor{red}{\boldmath$75.88 \pm 1.18$}  & $75.38 \pm 0.90$  & $73.27 \pm 0.50$  & $74.53 \pm 1.04$ \\
\cline{2-10}

& SortPool  & $72.90 \pm 0.61$   & $74.83 \pm 1.71$        & $73.52 \pm 0.53$        & $74.99 \pm 0.16$        & $74.56 \pm 0.39$  & \textcolor{red}{\boldmath$75.23 \pm 0.85$}  & $74.34 \pm 0.70$ & $75.08 \pm 0.53$ \\
\cline{2-10}

& EdgePool  & $73.03 \pm 0.96$   & $73.05 \pm 0.72$        & $73.98 \pm 0.54$        & $75.60 \pm 0.72$     & $74.19 \pm 0.53$  & $75.11 \pm 1.17$  & $74.56 \pm 1.80$ & \textcolor{red}{\boldmath$76.24 \pm 1.28$} \\
\cline{2-10}

& SagPool  & $74.71 \pm 1.09$   & $75.57 \pm 0.80$        & $73.52 \pm 0.80$        & $75.66 \pm 1.74$         &  $74.96 \pm 0.59$  & $74.53 \pm 0.54$  & \textcolor{red}{\boldmath$75.78 \pm 2.17$} & $74.86 \pm 1.32$ \\
\cline{2-10}

& GIN0  & $75.48 \pm 0.60$   &  $76.09 \pm 1.15$ & $75.26 \pm 1.95$        & $73.79 \pm 0.56$ &$75.14 \pm 0.59$  & \textcolor{red}{\boldmath$76.24 \pm 0.85$}  & $74.99 \pm 1.00$ & $74.44 \pm 0.62$  \\

\midrule
\multirow{6}{*}{RSVP}
& GraphSage  & $93.27 \pm 0.05$   & $93.25 \pm 0.35$        & $93.05 \pm 0.11$        & \textcolor{red}{\boldmath$93.47 \pm 0.06$}         & $93.22 \pm 0.08$  & $93.09 \pm 0.33$  & $93.08 \pm 0.20$ & $93.23 \pm 0.17$  \\
\cline{2-10}

& Set2Set    & $93.33 \pm 0.09$   & $93.19 \pm 0.22$        & $92.94 \pm 0.11$        & \textcolor{red}{\boldmath$93.34 \pm 0.17$}        & $93.24 \pm 0.26$  & $93.30 \pm 0.24$  & $93.32 \pm 0.10$ & $93.26 \pm 0.07$  \\
\cline{2-10}

& SortPool  & $93.24 \pm 0.20$   & $93.36 \pm 0.16$        & \textcolor{red}{\boldmath$93.39 \pm 0.28$}        & $93.38 \pm 0.27$        & $93.29 \pm 0.21$  & $93.05 \pm 0.34$  & $93.31 \pm 0.14$ & $93.35 \pm 0.24$ \\
\cline{2-10}

& EdgePool  & $92.89 \pm 0.04$   & $93.02 \pm 0.26$        & $93.32 \pm 0.06$        & $93.47 \pm 0.49$     & $93.39 \pm 0.23$  & $93.31 \pm 0.06$  & \textcolor{red}{\boldmath$93.49 \pm 0.17$} & $93.43 \pm 0.22$ \\
\cline{2-10}

& SagPool  & \textcolor{red}{\boldmath$93.45 \pm 0.19$}   & $93.34 \pm 0.09$        & $93.14 \pm 0.23$       & $92.99 \pm 0.16$         &  $93.36 \pm 0.02$  & $93.24 \pm 0.20$  & $93.03 \pm 0.07$ & $93.07 \pm 0.11$ \\
\cline{2-10}

& GIN0  & \textcolor{red}{\boldmath$93.26 \pm 0.07$}   & $93.23 \pm 0.01$        & $93.18 \pm 0.10$        & $93.07 \pm 0.19$         &  $93.23 \pm 0.11$  & $93.14 \pm 0.34$  & $93.10 \pm 0.18$ & $93.22 \pm 0.10$ \\

\bottomrule
\end{tabular}

\end{table*}
\begin{table*}[t]
\centering
\caption{Performance of datasets: Edge index construction using a distance threshold for the formation of edges, and distance threshold containing self-loops.}
\label{tab:dist}

\scriptsize
\begin{tabular}{c c c c c c c c}
\toprule
Dataset & Model &
\multicolumn{3}{c}{Distance} & 
\multicolumn{3}{c}{Distance w. Self-Loops}\\
\cmidrule(lr){3-5} \cmidrule(lr){6-8}  
& & d=0.3 & d=0.4 & d=0.5 & d=0.3 & d=0.4 & d=0.5 \\

\toprule
\multirow{6}{*}{ErrP}
& GraphSage  & $75.05 \pm 0.45$   & $74.86 \pm 1.50$        & \textcolor{red}{\boldmath$75.81 \pm 1.99$}        & $75.78 \pm 1.02$         &  $74.89 \pm 0.50$  & $74.89 \pm 0.61$  \\
\cline{2-8}

& Set2Set    & $74.68 \pm 2.02$   & $74.07 \pm 0.71$        & $75.02 \pm 0.57$        & $75.57 \pm 0.38$         &  \textcolor{red}{\boldmath$76.15 \pm 1.11$}  & $74.25 \pm 1.20$  \\
\cline{2-8}

& SortPool  & \textcolor{red}{\boldmath$75.94 \pm 0.95$}  & $74.16 \pm 0.26$        & $74.34 \pm 1.59$        & $74.99 \pm 1.62$         &  $73.79 \pm 0.84$  & $74.62 \pm 0.45$  \\ 
\cline{2-8}

& EdgePool  & $74.71 \pm 0.75$   & $74.83 \pm 0.41$        & $74.47 \pm 1.53$        & \textcolor{red}{\boldmath$75.84 \pm 0.55$}         &  $75.23 \pm 0.74$  & $74.56 \pm 1.23$  \\
\cline{2-8}

& SagPool  & $76.06 \pm 0.74$   & $74.40 \pm 1.32$        & \textcolor{red}{\boldmath$76.15 \pm 0.53$}        & $73.88 \pm 0.67$         &  $74.47 \pm 1.20$  & $74.80 \pm 1.28$  \\
\cline{2-8}

& GIN0  & \textcolor{red}{\boldmath$76.06 \pm 0.91$}   & $74.65 \pm 0.66$        & $74.83 \pm 0.85$        & $76.03 \pm 1.05$         &  $75.75 \pm 0.51$  & $75.97 \pm 0.46$  \\

\bottomrule
\end{tabular}

\end{table*}
\begin{table*}[t]
\centering
\caption{Performance of datasets: Hyperparameter selection for L1, L2, and ElasticNet Regularization of GNN models.}
\label{tab:res}

\scriptsize
\begin{tabular}{c c c c c c c c c}
\toprule
Dataset & Model &
\multicolumn{3}{c}{L1 Regularization} & \multicolumn{3}{c}{L2 Regularization} & \multicolumn{1}{c}{ElasticNet Regularization} \\
\cmidrule(lr){3-5} \cmidrule(lr){6-8}  
& & $\alpha=0.1$ & $\alpha=0.01$ & $\alpha=0.001$ & $\beta=0.2$ & $\beta=0.4$ & $\beta=0.8$ & Best of $\alpha \& \beta$\\

\toprule

\multirow{6}{*}{ErrP}
& GraphSage  & \textcolor{red}{\boldmath$76.55 \pm 0.87$}   & $76.30 \pm 1.00$        & $74.07 \pm 0.23$        & $75.54 \pm 0.44$        & $74.56 \pm 1.20$  & $74.93 \pm 1.41$  & $73.39\pm 0.22$ \\
\cline{2-9}

& Set2Set    & $74.86 \pm 0.20$   & $74.65 \pm 1.48$        & $74.56 \pm 1.65$        & $74.68 \pm 0.67$         & \textcolor{red}{\boldmath$76.18 \pm 0.19$}  & $74.22 \pm 0.96$  & $74.50\pm 0.34$ \\
\cline{2-9}

& SortPool & $75.14 \pm 0.33$ & $74.89 \pm 1.78$          & $74.80 \pm 1.74$        & $73.73 \pm 1.36$         & \textcolor{red}{\boldmath$75.26 \pm 0.71$}  & $74.65 \pm 0.74$  & $73.94\pm 0.82$ \\
\cline{2-9}

& EdgePool  & $73.64 \pm 0.17$ & $74.96 \pm 1.02$          & \textcolor{red}{\boldmath$76.15 \pm 1.06$}        & $74.99 \pm 0.46$         & $74.96 \pm 1.05$  & $75.23 \pm 0.23$  & $73.76\pm 0.67$ \\
\cline{2-9}

& SagPool  & $75.60 \pm 0.40$ & \textcolor{red}{\boldmath$76.58 \pm 0.54$}          & $75.11 \pm 0.30$        & $74.13 \pm 0.23$         & $74.19 \pm 1.13$  & $75.23 \pm 1.04$  & $73.58\pm 0.36$ \\
\cline{2-9}

& GIN0  & $75.60 \pm 0.42$ & $75.02 \pm 0.55$          & $75.17 \pm 1.01$        & $74.40 \pm 1.35$         & \textcolor{red}{\boldmath$76.73 \pm 0.40$}  & $74.56 \pm 1.06$  & $74.04\pm 0.85$ \\

\midrule
\multirow{6}{*}{RSVP}
& GraphSage  & $92.51 \pm 0.19$   & \textcolor{red}{\boldmath$93.49 \pm 0.10$}        & $93.07 \pm 0.32$        & $92.64 \pm 0.28$         & $92.60 \pm 0.11$  & $92.76 \pm 0.13$  & $92.89\pm 0.12$ \\
\cline{2-9}

& Set2Set    & $93.03 \pm 0.17$   & $93.12 \pm 0.13$        & \textcolor{red}{\boldmath$93.22 \pm 0.27$}        & $92.93 \pm 0.08$        & $92.97 \pm 0.11$  & $93.03 \pm 0.30$  & $91.47\pm 0.20$ \\
\cline{2-9}

& SortPool  & \textcolor{red}{\boldmath$93.14 \pm 0.12$}   & $93.11 \pm 0.20$       & $92.91 \pm 0.13$        & $92.71 \pm 0.31$         & $93.10 \pm 0.13$  & $92.74 \pm 0.08$  & $92.90\pm 0.12$ \\
\cline{2-9}

& EdgePool  & \textcolor{red}{\boldmath$93.49 \pm 0.10$}   & $93.29 \pm 0.10$        & $93.12 \pm 0.04$        & $93.12 \pm 0.01$         & $93.13 \pm 0.19$  & $92.74 \pm 0.19$  & $92.57\pm 0.16$ \\
\cline{2-9}

& SagPool  & $93.09 \pm 0.18$   & $93.18 \pm 0.21$       & \textcolor{red}{\boldmath$93.38 \pm 0.30$}        & $92.74 \pm 0.09$         & $93.08 \pm 0.17$  & $92.91 \pm 0.34$  & $92.87\pm 0.23$ \\
\cline{2-9}

& GIN0  & $92.87 \pm 0.10$  & \textcolor{red}{\boldmath$93.26 \pm 0.12$}        & $93.13 \pm 0.21$        & $92.93 \pm 0.06$         & $93.07 \pm 0.13$  & $92.49 \pm 0.13$  & $92.85\pm 0.15$ \\

\bottomrule

\end{tabular}

\end{table*}

%%%%%%%%%%%%%%%%%%%%%%%%%%%%%%%%%%%%%%%%%%%%%%%%%%%%%%%%%%%%%%%%%%%%%%%%%%%%%%%%

%%%%%%%%%%%%%%%%%%%%%%%%%%%%%%%%%%%%%%%%%%%%%%%%%%%%%%%%%%%%%%%%%%%%%%%%%%%%%%%%
\bibliographystyle{IEEEtran}
\bibliography{refs}

\end{document}